\pdfoutput=1
\documentclass{article}
\usepackage{spconf,amsmath,graphicx}
\usepackage{xcolor}
\usepackage{url}
\usepackage[hyperindex,breaklinks]{hyperref}
\usepackage{multirow}
\usepackage{makecell}
\usepackage[utf8]{inputenc}

\title{Comparing the Benefit of Synthetic Training Data for Various Automatic Speech Recognition Architectures}
\name{Nick Rossenbach$^{1,2}$, Mohammad Zeineldeen$^{1,2}$, Benedikt Hilmes$^{1}$, Ralf Schlüter$^{1,2}$, Hermann Ney$^{1,2}$}

\address{
  $^1$Human Language Technology and Pattern Recognition,
  Computer Science Department, RWTH Aachen University, 52074 Aachen, Germany \\
  $^2$AppTek GmbH, 52062 Aachen, Germany}
\begin{document}
%
\maketitle
\begin{abstract}
Recent publications on automatic-speech-recognition (ASR) have a strong focus on attention encoder-decoder (AED) architectures which tend to suffer from over-fitting in low resource scenarios.
One solution to tackle this issue is to generate synthetic data with a trained text-to-speech system (TTS) if additional text is available.
This was successfully applied in many publications with AED systems, but only very limited in the context of other ASR architectures.
We investigate the effect of varying pre-processing, the speaker embedding and input encoding of the TTS system w.r.t.\ the effectiveness of the synthesized data for AED-ASR training.
Additionally, we also consider internal language model subtraction for the first time, resulting in up to 38\% relative improvement.
We compare the AED results to a state-of-the-art hybrid ASR system, a monophone based system using connectionist-temporal-classification (CTC) and a monotonic transducer based system.
We show that for the later systems the addition of synthetic data has no relevant effect, but they still outperform the AED systems on LibriSpeech-100h.
We achieve a final word-error-rate of 3.3\%/10.0\% with a hybrid system on the clean/noisy test-sets, surpassing any previous state-of-the-art systems on Librispeech-100h that do not include unlabeled audio data.
\end{abstract}
\section{Introduction}
\label{sec:intro}
Many publications \cite{Synnaeve-2019-End-to-endASRfrom, park2019specaugment, Karita-2019-AComparativeStudy} have shown that ASR systems using an AED architecture can achieve similar performance on large corpora  compared to other methods such as hidden-markov-model (HMM) based hybrid models \cite{Luscher-2019-RWTHASRSystemsfor}, CTC models \cite{Graves06connectionisttemporal, pmlr-v32-graves14} or recurrent neural network Transducer (RNN-T) models \cite{DBLP:journals/corr/abs-1211-3711, Han-2020-ContextNetImprovin}.
For smaller corpora however, they usually suffer from a much stronger performance loss \cite{Luscher-2019-RWTHASRSystemsfor,Park-2020-ImprovedNoisyStude}. To increase the effectiveness of attention-based ASR systems different methods were proposed, such as data augmentation techniques like SpecAugment \cite{DBLP:journals/corr/abs-1904-08779}, various regularization techniques \cite{Tske2020SingleHA}, generating synthetic data from additional text \cite{Baskar-2019-Semi-SupervisedSequ, Rosenberg-2019-SpeechRecognitionw, Rossenbach-2020-GeneratingSynthetic,Laptev-2020-YouDoNotNeedMore, Baskar-2021-EatEnhancedASR-TT} or using unlabeled speech data \cite{Park-2020-ImprovedNoisyStude, Baskar-2021-EatEnhancedASR-TT, Kahn-2020-Self-TrainingforEn}.

In this work we will make use of SpecAugment and a Tacotron-2 \cite{Shen-2018-NaturalTTSSynthesi} style TTS system to boost the performance of ASR systems on LibriSpeech-100h \cite{libirispeech}, which is a quite common task to test the performance of synthetic data or semi-supervised training approaches. LibriSpeech-360h and LibriSpeech-500h are then used as text-only or audio-only data.
In this context we experiment with different text encoding, speaker encoding and a novel approach to improve on the stability issues in autoregressive TTS systems.
These issues are caused by the nature of the ASR data in contrast to TTS targeted data \cite{Rosenberg-2019-SpeechRecognitionw, Zen-2019-LibriTTSACorpusD}. It was shown in \cite{Zen-2019-LibriTTSACorpusD} that the stability does not only depend on the used TTS system, but also on the quality and processing style of the data.
LibriSpeech contains many utterances with unnaturally long pauses, which are causing stability issues in TTS, also referred to as "bias-problem" \cite{Liu-2020-Teacher-StudentTrai}.
While it is impossible to have an objective metric for TTS quality without using human ratings, \cite{Shen-2020-Non-AttentiveTacotr} presented two objective metrics which give an indication for stability issues.

Most previous publications on generating synthetic data for LibriSpeech-100h only aimed at improving a single AED system \cite{Baskar-2019-Semi-SupervisedSequ,Rossenbach-2020-GeneratingSynthetic,Laptev-2020-YouDoNotNeedMore}. In this work, we will use four fundamentally different architectures to show the effects of synthetic data. For RNN-T systems, there is prior work on domain adaptation with TTS \cite{Li-2020-DevelopingRNN-TMod} and a recent publication on improving LibriSpeech \cite{Fazel-2021-SynthASRUnlocking} by using synthetic data. The four state-of-the-art baselines we present are an AED system, a hybrid ASR system, a CTC system and a monotonic RNN-T system.
One objective of this paper is to determine how much we can close the gap between an AED system and a hybrid system by adding synthetic data under fair conditions.
To the best of our knowledge no recent publication shows strong results with a hybrid ASR system on LibriSpeech-100h, with the previous best results being reported in \cite{Luscher-2019-RWTHASRSystemsfor}. We wanted to see if models without label context such as the hybrid and the CTC acoustic models can improve with synthetic data, and if yes,  how much. In \cite{Rossenbach-2020-GeneratingSynthetic} it was shown that synthetic data and language model fusion have orthogonal effects. As ASR systems with label context such as AED and RNN-T benefit from subtracting an estimated internal language model (ILM) \cite{9003790, Variani-2020-HybridAutoregressiv, Meng2021InternalLM, Zeineldeen-2021-InvestigatingMethod, DBLP:journals/corr/abs-2104-03006}, we investigate if the benefits of synthetic data from additional text are retained when using ILM subtraction methods for AED systems.

%
We make general versions of the used RETURNN\cite{DBLP:conf/icassp/DoetschZVKSN17} and RASR\cite{Wiesler-2014-RASRNNTheRWTHne} toolkit configuration files available online\footnote{\scriptsize\url{https://github.com/rwth-i6/returnn-experiments/tree/master/2021-tts-asr-librispeech100h}}.
All experiments in this work were managed with Sisyphus \cite{DBLP:conf/emnlp/PeterBN18}.

\section{Speech Recognition}

\subsection{Attention-based ASR}

The AED system is inspired by previous work on attention-based ASR systems using bidirectional long-short-term-memory (BLSTM)\cite{hochreiter1997lstm} layers as encoder and decoder 
\cite{Zeyer-2018-ImprovedTrainingof, zeyer2019:trafo-vs-lstm-asr}.
Our encoder consists of 2 convolutional layers followed by 6 BLSTM layers with 1024 dimensions per direction.
We use time downsampling via max-pooling layers with a factor of 6.
We apply regularization techniques for both encoder and decoder similar to \cite{Tske2020SingleHA}.
We apply a dropout \cite{10.5555/2627435.2670313} of $30\%$ to the LSTM input and $30\%$ drop-connect to the
recurrent hidden-to-hidden weight matrices.
Our decoder consists of a Zoneout-LSTM layer \cite{Krueger-2016-ZoneoutRegularizin} with a dimension of 1000.
We apply $30\%$ attention dropout and embedding dropout, as well as using $10^{-3}$ weight decay and $0.1$ label smoothing.
We use byte-pair-encoding (BPE) \cite{sennrich-etal-2016-neural} as output labels with a vocabulary size of 2k.
We add CTC as additional
loss on top of the encoder for training stability.

\subsection{Hybrid ASR}
\label{sec:hybrid}

The hybrid ASR system used in this paper follows the train-clean-100 model presented in \cite{Luscher-2019-RWTHASRSystemsfor}.
The initial alignment is created by using a Gaussian-mixture-model (GMM) that is as described in Section 2.1.1 of the aforementioned paper.
The hybrid BLSTM acoustic model (AM) consists of 8 BLSTM-layers with 1024 hidden units per direction. The model uses 12k classification and regression tree (CART) \cite{young1992cart} based labels and is trained with standard cross-entropy loss. We do not use dropout or L2 constraints for the hybrid training. The BLSTM network is implemented in RETURNN, the GMM-training and the decoder in RASR.

\subsection{CTC ASR}

Similar to the monotonic RNN-T model presented in \cite{Zhou-2021-PhonemeBasedNeural}, our CTC model uses a 6-Layer BLSTM network, with a factor 2 max-pooling layer applied between the third and fourth layer.
Each LSTM layer has a hidden dimension of 512 per direction and includes an L2 loss and dropout of 0.1 for regularization.
The CTC labels consists of the CMUDict\footnote{\scriptsize\url{http://www.speech.cs.cmu.edu/cgi-bin/cmudict/}} phonemes with and without an additional end-of-word labeling symbol "\#", and a unified silence/blank symbol.
This results in a total number of $139$ labels. The BLSTM network is implemented in RETURNN. The state machine for Baum-Welch loss computation used during training and the decoder are implemented in RASR.

\subsection{RNN-Transducer ASR}

The monotonic RNN-T model also follows \cite{Zhou-2021-PhonemeBasedNeural}, using the same encoder model and labels as the CTC model. The prediction network is a 2-layer LSTM with 1024 dimensions each, so other than in \cite{Zhou-2021-PhonemeBasedNeural} the model uses the complete history for the prediction network. During training three different losses are used. First, the model includes a CTC loss on the encoder output states which is the same as in the CTC model. Then, there is both a normal CE loss with the aligned labels as target, but also a segmental loss which excludes blank positions.

\subsection{Language Modeling}

For the attention-based setup we trained a 24-layer Transformer language model (LM) following the setup described in \cite{Irie-2019-LanguageModelingwi}.
The BPE-perplexity of the final model on the dev-sets is 18.6, which would be equivalent to 73.6 on word-level.
For the hybrid and CTC setup, we use a 2-layer LSTM network with a hidden dimension of 4096 and an output vocabulary of 200k, as a Transformer network is computationally too expensive in these scenarios.
The word-level perplexity on the dev-sets is 60.

\section{Speech Synthesis}
\vspace{-0.05em}
The synthesis system consists of an LSTM-based encoder-decoder architecture with a location-sensitive attention.
This architecture is commonly referred to as "Tacotron-2" \cite{Shen-2018-NaturalTTSSynthesi}.
The setup closely follows \cite{Rossenbach-2020-GeneratingSynthetic}, with a different variation of input and speaker embedding as well as minor changes.
The Zoneout-LSTM in the encoder was replaced by our native CUDA LSTM implementation and a dropout layer for increased training and decoding speed.
For the decoder, we kept the 2-layer Zoneout-LSTMs, but used a hidden dimension of 768.
To convert the log-mel features of the TTS system into audio-signals, we use a separately trained 2-layer BLSTM to convert the features first back into linear STFT features \cite{Rossenbach-2020-GeneratingSynthetic}.
The linear features are transformed into raw-waveforms with the Griffin \& Lim \cite{DBLP:conf/icassp/GriffinDL84} algorithm  and stored as .ogg files.
We found out that the usage of .ogg files has no effect on the ASR training. Also, the synthetic data is required to be stored in audio waveform instead of features, as the different ASR systems have independent and varying feature pipelines.
\vspace{-0.05em}
\subsection{Input Encoding}
\vspace{-0.05em}
In this work we experimented with character embeddings as well as phoneme embeddings.
The phoneme representations are based on the CMU pronunciation dictionary. We use Sequitur \cite{Bisani-2008-Joint-sequencemodel} to generate the phoneme representations for words not part of the dictionary.

\subsection{Silence Preprocessing}
\label{sec:preprocessing}

The LibriSpeech corpus has an utterance structure that is not beneficial for TTS systems \cite{Zen-2019-LibriTTSACorpusD}. Especially longer silence parts that occur within an utterance can severely influence the model performance.
In this paper we compare two approaches for silence pre-processing, namely threshold-based silence filtering and a newly introduced GMM alignment-based silence filtering.
For threshold-based filtering we use the FFMPEG silence-remove filter with a (silence-)threshold value of -40dB. For the GMM-based silence processing we train a simple GMM-HMM model starting from a linear alignment.
It uses 16-dimensional MFCC features with first and second derivatives as well as energy features, as in \cite{Luscher-2019-RWTHASRSystemsfor}. The alignment and the mixture densities are updated over 75 iterations, followed by 10 iterations splitting and re-estimating the mixture densities.
A final alignment is used to determine all silence frames. We extract timestamps for all silence regions, and remove those parts excluding a pre-defined amount $\Delta t$ at the silence region borders from the audio files. This means for a silence region from, e.g., $2s$ to $4s$ in an utterance we would remove the audio from $2.25s$ to $3.75s$ if $\Delta t = 0.5s$, or all of it if $\Delta t = 0s$.
If any, we keep only silence between words, not before or after the utterance.

\subsection{Speaker Modeling}

For controlling the speaker characteristics of the TTS system, we compare a supervised and an unsupervised approach.
For the supervised speaker embeddings we simply add a 256-dimensional look-up table for all 251 speakers of the training corpus.
As unsupervised method we use a reference encoder with global style tokens (GST) \cite{Wang-2018-StyleTokensUnsupe}.
The reference encoder and the GST network are implemented as described in \cite{Rossenbach-2020-GeneratingSynthetic}.

\section{Experiments}

All our experiments are conducted by using LibriSpeech-100h as training data for the TTS and the baseline ASR systems, and the text of LibriSpeech-360h for synthesizing additional training data.
We choose this scenario, as this is a very common task for semi-supervised training, and allows to compare our results with previous literature \cite{Baskar-2019-Semi-SupervisedSequ, Rossenbach-2020-GeneratingSynthetic, Laptev-2020-YouDoNotNeedMore, Baskar-2021-EatEnhancedASR-TT}.
The best checkpoints are determined by the minimum of the negative log-likelihood score on a holdout set consisting of a subset of dev-clean and dev-other. All systems use SpecAugment \cite{park2019specaugment} in training. Note that we optimized the parameters and training settings for each of the different ASR architectures presented individually in order to achieve the respective best performance on LibriSpeech-100h. The scales for language model integration were tuned on dev-clean/dev-other.

\subsection{Evaluation Methods}

As evaluation method for our ASR systems we use word error rate (WER).
For the evaluation of our TTS systems we use two metrics as proposed in \cite{Shen-2020-Non-AttentiveTacotr}: Word Deletion Rate (WDR) and Unaligned Duration Ratio (UDR).
The WDR is defined as the relative amount of non-generated words by the TTS, which is the deletion rate in WER evaluation.
The UDR is given by the ratio of not aligned audio segments of a length greater than a certain threshold, which was defined to be one second.
For further details on these metrics refer to \cite{Shen-2020-Non-AttentiveTacotr}.

\begin{table}[htb]
\caption{UDR and WDR measured on dev-clean and test-clean when synthesized with TTS models using the respective preprocessed data. Values are averaged over 4 different trained models each.}
\label{tab:silence}
\begin{center}
\setlength{\tabcolsep}{5pt}
\begin{tabular}{|c|c|c|c|}
\hline
\multicolumn{2}{|c|}{\multirow{2}{*}{\makecell{Silence\\Pre-Proc.}}} & UDR $[\%]$&WDR$[\%]$\\
\cline{3-4}
\multicolumn{2}{|c|}{} & \multicolumn{2}{c|}{dev + test}\\
\cline{3-4}
& $\Delta t[s]$ & \multicolumn{2}{c|}{clean}\\
\hline
\hline
\multicolumn{2}{|c|}{Threshold} & 9.8 & 3.5\\
\hline
GMM & 0.0 & 0.0 & 2.9\\
\hline
GMM & 0.5 & 0.0 & 3.0\\
\hline
\end{tabular}
\end{center}
\end{table}

\begin{table}[t]
\caption{Results on LibriSpeech-100h with an AED model without external LM. The TTS models for data generation used a fixed look-up table and phoneme inputs.}
\label{tab:AED}
\begin{center}
\setlength{\tabcolsep}{5pt}
\begin{tabular}{|l|c|c|c|c|c|c|c|}
\hline
\multirow{3}{*}{\makecell{Syn.\\Data}} & \multicolumn{2}{c|}{\multirow{2}{*}{\makecell{Silence\\Pre-Proc.}}} & \multirow{2}{*}{\makecell{Added\\Data}} & \multicolumn{4}{c|}{WER$[\%]$}\\
\cline{5-8}
 & \multicolumn{2}{c|}{} & & \multicolumn{2}{c|}{dev} &\multicolumn{2}{c|}{test} \\
\cline{5-8}
 & & $\Delta t[s]$ & $[h]$ &cl.&oth.&cl.&oth.\\
\hline
\hline
No & \multicolumn{2}{c|}{-} & - &8.1&21.6&8.2&22.6\\
\hline
\multirow{3}{*}{Yes} & \multicolumn{2}{c|}{Threshold}&330 &5.7&19.7&6.0&20.9\\
\cline{2-8}
 & \multicolumn{1}{l|}{\multirow{2}{*}{GMM}} & 0.5 & 345 &5.6&19.8&6.1&20.5\\
\cline{3-8}
 & & 0.0 & 278 &5.7&19.5&5.9&20.3\\
\hline
Oracle & \multicolumn{2}{c|}{-} & 360 & 4.5 & 14.8 & 4.8 & 15.3\\
\hline
\end{tabular}
\end{center}
\end{table}
\vspace{-1.0em}
\subsection{Stability Analysis of the TTS}
\label{sec:stability}

For each trained TTS system, we synthesized the text of LibriSpeech dev-clean and test-clean and computed the UDR and WDR using existing ASR systems.
For the UDR, we used a GMM-HMM ASR model and ran a forced-alignment on the synthesized data to label each frame as "speech" or "silence/noise".
Following \cite{Shen-2020-Non-AttentiveTacotr}, we use one second as threshold for allowed non-aligned-audio.
To compute the WDR, we use an attention-based LibriSpeech-960h ASR system \cite{Luscher-2019-RWTHASRSystemsfor}, with a WDR on dev-clean and test-clean of 0.5\%.
In the first set of experiments we trained 12 different TTS systems by adjusting 3 different conditions:
\vspace{-0.6em}
\begin{itemize}
\setlength{\itemindent}{-.1in}
    \item Input encoding:
    \begin{itemize}
        \vspace{-0.7em}
        \setlength{\itemindent}{-.25in}
        \item Lower cased characters (\textbf{char})
        \vspace{-0.4em}
        \item CMUDict-style phonemes with stress (\textbf{phon})
    \end{itemize}
    \vspace{-0.6em}
    \item Speaker Encoding
    \begin{itemize}
        \vspace{-0.7em}
        \setlength{\itemindent}{-.25in}
        \item Trained look-up table (\textbf{look-up})
        \vspace{-0.4em}
        \item Reference encoder with Global-Style-Token (\textbf{GST})
    \end{itemize}
    \vspace{-0.6em}
    \item Silence Pre-Processing
    \begin{itemize}
        \vspace{-0.7em}
        \setlength{\itemindent}{-.25in}
        \item Threshold-based with -40dB (\textbf{Threshold})
        \vspace{-0.4em}
        \item GMM-HMM alignment w. 0.0s silence (\textbf{GMM 0.0s})
        \vspace{-0.4em}
        \item GMM-HMM alignment w. 0.5s silence (\textbf{GMM 0.5s})
    \end{itemize}
\end{itemize}
First, we used the UDR to determine if using the GMM-HMM alignment to remove excessive silence helps to reduce long noise and silence sections in the synthesized audio.
Table \ref{tab:silence} shows the results with UDR and WDR for each silence pre-processing method, averaged over the remaining conditions (speaker and input encoding). By using the GMM-HMM alignment for silence pre-processing, the UDR can be reduced to an absolute zero, so there are no unaligned noise or silence parts above one second remaining in the synthesized audio. This also means that the TTS system always stops correctly at the end of an utterance. The WDR is reduced by up to 0.6\%, meaning fewer words are dropped during synthesis, but is still higher than the original 0.5\%. This indicates that the TTS still suffers from either early stopping or skipped words.

\begin{table}[htb]
\caption{Comparison of different TTS systems. The WER score is the AED system performance after synthesizing LibriSpeech-360h and using it for training.
* means averaged over all possible TTS models and respective ASR training.}
\label{tab:input_conditions}
\begin{center}
\setlength{\tabcolsep}{5pt}
\begin{tabular}{|c|c|c|c|c|c|c|}
\hline
\multicolumn{2}{|c|}{\multirow{2}{*}{\makecell{Silence\\Pre-Proc.}}} & \multirow{3}{*}{Speaker} & \multirow{3}{*}{\makecell{Input\\Type}}&WDR$[\%]$&\multicolumn{2}{c|}{WER$[\%]$}\\
\cline{5-7}
\multicolumn{2}{|c|}{} & & & dev + test & \multicolumn{2}{c|}{test}\\
\cline{5-7}
& $\Delta t[s]$ & & & clean & cl. & oth. \\
\hline
\hline
\multicolumn{2}{|c|}{Threshold} & * & *  & 3.5 & 6.2 & 20.9 \\
\hline
GMM &0.0& * & * & 2.9 & 6.1 & 20.8 \\
\hline
GMM &0.5& * & * & 3.0 & 6.2 & 20.8 \\
\hline
\hline
\multicolumn{2}{|c|}{*}& look-up & * & 3.0 & 6.1 & 20.5 \\
\hline
\multicolumn{2}{|c|}{*}& GST & * & 3.6 & 6.2 & 21.1 \\
\hline
\hline
\multicolumn{2}{|c|}{*}& * & char & 3.2 & 6.2 & 20.8 \\
\hline
\multicolumn{2}{|c|}{*}& * & phon & 3.1 & 6.1 & 20.8 \\
\hline
\end{tabular}
\end{center}
\end{table}

\begin{table}[htb]
\caption{Comparing the effect of different data ratios of librispeech-100h compared to the synthesized LibriSpeech-360h corpus (real \textnormal{:} synthetic), assuming that the synthesized corpus has a duration of about 300h. The results are the average of 4 experiments with the TTS systems used as stated.}
\label{tab:data_ratio}
\begin{center}
\begin{tabular}{|c|c|c|c|c|c|}
\hline
\multirow{3}{*}{\makecell{Data\\Ratio}} & \multirow{3}{*}{\makecell{Silence\\Pre-Proc.}} & \multirow{3}{*}{Speaker} & \multirow{3}{*}{\makecell{Input\\Type}} &\multicolumn{2}{c|}{WER$[\%]$}\\
\cline{5-6}
 & & & &\multicolumn{2}{c|}{test} \\
\cline{5-6}
 & & & & cl. & oth. \\
\hline
\hline
1:3 &  \multirow{3}{*}{\makecell{GMM 0.0/\\0.5}} &  \multirow{3}{*}{GST/look-up} & \multirow{3}{*}{phon} & 6.3 &  22.4\\
\cline{1-1}
\cline{5-6}
2:3 & & & & 6.0 & 21.1\\
\cline{1-1}
\cline{5-6}
3:3 & & & & 6.1 & 20.7\\
\hline
\end{tabular}
\end{center}
\end{table}

\subsection{Synthetic data for Attention-Encoder-Decoder ASR}
\label{sec:aed}

As AED systems use less resources in training and decoding we first tested the quality of the different TTS systems with our AED-ASR baseline.
The baseline is trained for 250 sub-epochs with a partitioning factor of 3, resulting in about 83 full epochs.
We reset the learning rate to the maximum at sub-epoch 80 for all experiments. When adding synthetic data, we take the checkpoint of sub-epoch 80 of the baseline as starting point to reduce the training variance for the experiments \cite{Rossenbach-2020-GeneratingSynthetic}.
For the remaining 170 sub-epochs we oversample the real data 3 times, having $3\cdot100$h of real data matched with $\sim$300h of synthetic data. With a partitioning of factor 9 this results in 83 epochs of training on LibriSpeech-100h, and about 19 epochs for the synthetic data.
The results for the baseline, synthetic data from 3 different TTS systems, and the oracle performance can be found in Table \ref{tab:AED}. We see a notable jump in WER of about 2\% absolute when adding the synthetic training data. The term oracle refers to using the original LibriSpeech-360h audio files instead of the synthesized one.
To see the effect of the different TTS systems better, we also combined all important conditions and created synthetic data. The evaluation of the different conditions can be seen in Table \ref{tab:input_conditions}.
For each condition, we average over all possible variants for the other conditions, i.e. for the first line we took the average of the 4 experiments with threshold silence pre-processing combined with look-up or GST embedding and character or phoneme encoding.
The results indicate that only the speaker embedding has a notable effect on the final ASR performance, although we observe an increased stability by using the GMM silence pre-processing method.
We conclude that the ASR does not suffer from noisy or incorrect sequences, but instead they have a regularization effect on the training.
For those 12 experiments, we picked 4 to also test different data ratios 1:3 and 2:3, which can be found in Table \ref{tab:data_ratio}. Here we confirmed that the AED model with oversampling LibriSpeech-100h by factor 3 is performing best.


\subsubsection{Internal Language Model Subtraction}

Internal language model (ILM) subtraction is a method to remove the label prediction bias of ASR models that use a label prediction context such as AED systems and RNN-T systems.
The first methods were presented in \cite{9003790, Variani-2020-HybridAutoregressiv, Meng2021InternalLM} and extensively investigated for AED systems in \cite{Zeineldeen-2021-InvestigatingMethod}.
Although other variants of ILM estimation can perform better, we use the "Zero" ILM estimation approach \cite{Zeineldeen-2021-InvestigatingMethod} for simplicity, meaning that the acoustic context vector is set to zero for the decoder to estimate the ILM.
Table \ref{tab:ILM} shows the results for the AED system when adding synthetic data to systems with and without ILM subtraction.
We see that for both test sets the absolute improvement by adding synthetic data stays the same, being 2.1\%/2.6\% without ILM subtraction and 2.0\%/2.4\% with ILM subtraction for clean/other respectively.
This means that the relative improvement is even larger, reaching 38\% on test-clean and 16\% on test-other.
\vspace{-1.0em}

\begin{table}[t]
\caption{Results on LibriSpeech-100h with an AED model without LM and with external LM and internal language model subtraction using the zero method.
The experiments with synthetic data are averaged over 3 runs with the respective silence pre-processing methods.}
\label{tab:ILM}
\begin{center}
\vspace{-0.2em}
\setlength{\tabcolsep}{5pt}
\begin{tabular}{|c|c|c|c|c|c|c|}
\hline
\multirow{3}{*}{\makecell{External LM}} & \multirow{3}{*}{\makecell{ILM\\subtraction}} & \multirow{3}{*}{\makecell{Syn.\\Data}} & \multicolumn{4}{c|}{WER$[\%]$}\\
\cline{4-7}
 & & & \multicolumn{2}{c|}{dev} &\multicolumn{2}{c|}{test} \\
\cline{4-7}
 & & &cl.&oth.&cl.&oth.\\
 \hline
 \hline
 - & \multirow{3}{*}{No} & \multirow{2}{*}{No} &8.1&21.6&8.2&22.6\\
 \cline{1-1}
 \cline{4-7}
 \multirow{4}{*}{Transformer} & & & 5.7 & 15.8 & 6.1 & 16.9\\
 \cline{3-7}
 & & Yes & 3.6 & 13.9 & 4.0 & 14.3\\
 \cline{2-2}
 \cline{3-7}
 & \multirow{2}{*}{Yes} & No & 4.6 & 13.7 & 5.3 & 14.8\\
 \cline{3-7}
 & & Yes & 3.0 & 12.0 & 3.3 & 12.4\\
\hline
\end{tabular}
\vspace{-1.4em}
\end{center}
\end{table}

\begin{table*}[t]
\caption{Comparison of different systems on LibriSpeech-100h from literature and the results for each architecture described in this paper. Synthetic data describes the TTS used to generate the synthetic data, which is always based on the transcriptions of LibriSpeech-360h. Except for \cite{Luscher-2019-RWTHASRSystemsfor}, all systems include SpecAugment.}

\label{tab:comparison}
\begin{center}
\resizebox{.95\textwidth}{!}
{
\begin{tabular}{|c|c|c|c|c|c|c|c|c|}
\hline
\multirow{3}{*}{Architecture} & \multirow{3}{*}{\makecell{Encoder Model}} & \multirow{3}{*}{Label Type} & \multirow{3}{*}{LM}  & \multirow{3}{*}{Synthetic Data}& \multicolumn{4}{c|}{WER$[\%]$}\\
\cline{6-9}
 & & & & & \multicolumn{2}{c|}{dev} &\multicolumn{2}{c|}{test} \\
\cline{6-9}
 & & & & & clean & other & clean & other \\
\hline
\hline
 \multirow{10}{*}{\makecell{Attention\\Enc.-Dec.}} & LAS - BLSTM \cite{Park-2020-ImprovedNoisyStude} & 16k WPM & - & - & 5.3 & 16.5 & 5.5 & 16.9 \\
\cline{2-9}
 &  \multirow{3}{*}{ESPNet - Transformer \cite{Laptev-2020-YouDoNotNeedMore}} & \multirow{3}{*}{\makecell{5k SPM}} & - & \multirow{2}{*}{-} & 10.3 & 24.0 & 11.2 & 24.9\\
\cline{4-4}
\cline{6-9}
 & & & \multirow{2}{*}{LSTM} & & 5.8 & 16.6 & 7.0 & 17.0 \\
 \cline{5-9}
 & & & & VAE-TTS & 3.8 & 13.2 & 4.3 & 13.5 \\
\cline{2-9}
& \multirow{3}{*}{ESPNet - Transformer \cite{Baskar-2021-EatEnhancedASR-TT}} & \multirow{3}{*}{characters} & \multirow{3}{*}{LSTM} & - & 14.3 & 36.4 & 14.4 & 36.9\\
\cline{5-9}
& & & & x-vector TTS & 8.9 & 23.0 & 8.6 & 24.1\\
\cline{5-9}
& & & & x-vector TTS + $\alpha$ & 4.5 & 15.8 & 4.7 & 15.9\\
\cline{2-9}
 & \multirow{3}{*}{RETURNN - BLSTM [ours]} & \multirow{3}{*}{2k BPE} & - & \multirow{2}{*}{-} & 8.1 & 21.7 & 8.2 & 22.6 \\
\cline{4-4}
\cline{6-9}
 & & & \multirow{2}{*}{\makecell{Transformer\\ + ILM sub.}} & & 4.6 & 13.7 & 5.3 & 14.8\\
  \cline{5-9}
 & & & & lookup-TTS$^*$& \textbf{3.0} & \textbf{12.0} & \textbf{3.3} & \textbf{12.4}\\
\hline
\hline
\multirow{4}{*}{Hybrid} & RETURNN - BLSTM \cite{Luscher-2019-RWTHASRSystemsfor} & \multirow{4}{*}{12k CART} &\multirow{2}{*}{4-gram} & \multirow{3}{*}{-} & 5.0 & 19.5 & 5.8 & 18.6\\
\cline{2-2}
\cline{6-9}
 & \multirow{3}{*}{RETURNN - BLSTM [ours]} & &  &  & 4.9 & 14.7 & 5.6 & 15.3\\
\cline{4-4}
\cline{6-9}
 & & & \multirow{2}{*}{LSTM} &  & \textbf{3.0} & \phantom{0}\textbf{9.3} & 3.4 & \textbf{10.0} \\
 \cline{5-9}
 & & & & lookup-TTS$^*$ & \textbf{3.0} & \phantom{0}\textbf{9.3} & \textbf{3.3} & \textbf{10.0} \\
\hline
\hline
\multirow{4}{*}{CTC} & wav2letter++ - CTC \cite{Likhomanenko-2020-slimIPLLanguage-Mo} & characters & Transformer & - & 3.3 & \textbf{10.8} & 3.8 & \textbf{11.3}\\
\cline{2-9}
 & \multirow{3}{*}{RETURNN - BLSTM[ours]} & \multirow{3}{*}{phonemes (w. EOW)} & 4-gram & \multirow{2}{*}{-} & 5.0 & 15.4 & 5.6 & 16.1\\
\cline{4-4}
\cline{6-9}
 & & & \multirow{2}{*}{LSTM} &  & 3.3 & 11.4 & 3.8 & 12.4 \\
 \cline{5-9}
 & & & & lookup-TTS$^*$ & \textbf{3.2} & 11.6 & \textbf{3.6} & 12.5 \\
\hline
\hline
\multirow{3}{*}{\makecell{monotonic\\RNN-T}} & \multirow{3}{*}{RETURNN - BLSTM[ours]} & \multirow{3}{*}{phonemes (w. EOW)} & 4-gram & \multirow{2}{*}{-} & 5.0 & 14.9 & 5.4 & 15.6 \\
\cline{4-4}
\cline{6-9}
 & & & \multirow{2}{*}{LSTM} &  & 3.3 & 10.4 & \textbf{3.6} & \textbf{11.0} \\
 \cline{5-9}
 & & & & lookup-TTS$^*$& \textbf{3.1} & \textbf{10.1} & \textbf{3.6} & 11.1\\
\hline
\end{tabular}}\\
\vspace{0.5em}
* \textit{results are averaged over multiple runs with varying silence pre-processing, see Section \ref{sec:stability}}
\vspace{-0.5em}
\end{center}
\end{table*}

\subsection{Synthetic Data for Hybrid ASR}

As hybrid systems require a given alignment, we cannot simply add the synthetic data to the training but need to compute an alignment for the synthetic data as well.
We decided to re-align both LibriSpeech-100h and the added synthetic data with the acoustic BLSTM-AM model of the baseline (see Section \ref{sec:hybrid}) and do a full re-training.
The second block of Table \ref{tab:comparison} shows the baseline with the BLSTM alignment and re-training, and the same training including synthetic data from the respective TTS model. We again oversample the real data 3 times, and train for 5 full epochs in total. Although there is an improvement in the cross-entropy scores on the dev-sets during training from 0.298 to 0.275, there is no notable improvement in WER for any kind of used synthetic data, even after re-tuning the LM scales.


\subsection{Synthetic Data for CTC ASR}

The baseline CTC system is trained for 300 sub-epochs with a partitioning of 3, resulting in 100 training epochs.
When adding the synthetic data the system was trained for 400 sub-epochs with a partitioning of 12.
Block 3 of Table \ref{tab:comparison} shows the results for decoding. Although the CTC loss on the dev-set during training was noticeably lower (0.8 to 0.73) the final WER did not improve noticeably.


\vspace{-1.0em}
\subsection{Synthetic Data for monotonic RNN-T ASR}

The monotonic RNN-T ASR system is trained using the Viterbi-alignment generated with the respective CTC system. Both the baseline and the models including synthetic data were trained for 300 sub-epochs with a partitioning of 3 and 12
 respectively. The results can be found in the last block of Table \ref{tab:comparison}. It can be seen that on the "other" sets the RNN-T system which includes a prediction network is stronger than the equivalent CTC model. Nevertheless, it does not seem to improve substantially by adding synthetic data.
\vspace{-0.2em}
\section{Discussion}

To put the performance of our systems into perspective, we show a full comparison of the best systems of each category in Table \ref{tab:comparison}.
Only the AED system improved significantly by using synthesized data from additional text, which is in line with previous publications that only show improvements for AED systems.
One exception is \cite{Fazel-2021-SynthASRUnlocking} which uses an RNN-T ASR system reporting up to 12.5\% improvements on LibriSpeech, but the TTS system includes additional training data which means the TTS can learn additional acoustic information.
In our case, when relying only on the LibriSpeech-100h dataset as parallel data, we had no improvements with the two label context-independent models (CTC and hybrid ASR) and the context-dependent RNN-T.
This indicates that the TTS system cannot produce a larger variety of audio features compared to the existing data. We observed in other experiments that AED systems did not improve when synthesizing exactly the same text that is already used for the baseline training.
Similar behavior was also found in \cite{Rosenberg-2019-SpeechRecognitionw}, and it was shown in \cite{Rossenbach-2020-GeneratingSynthetic, Baskar-2021-EatEnhancedASR-TT} that training with synthetic data is complementary to using SpecAugment.
Our conclusion was that the improvements for AED systems shown in many publications are due to the improved decoder. The large improvements by reducing the influence of encoder states based on synthetic data, as done with the $\alpha$-factor in \cite{Baskar-2021-EatEnhancedASR-TT} further backed this hypothesis.
Now that we see equivalent improvements when using ILM subtraction for AED-ASR and no improvements with RNN-T models, we conclude that it is not the internal language model of the decoder that benefits most from the synthetic data, but rather the attention mechanism. It was also previously shown in \cite{Rossenbach-2020-GeneratingSynthetic} that the effect of synthetic data from additional text and external language model fusion of the same text have independent effects. Another possibility is that RNN-T systems might not need a large label context \cite{Zhou-2021-PhonemeBasedNeural, 9054419}, and thus cannot benefit from more textual data. Investigating if the label type plays a role in the effect of synthetic data can be future work.\\
\\
We also compare our systems to other publications.
The system from \cite{Park-2020-ImprovedNoisyStude} is to our knowledge the best existing LibriSpeech-100h AED system without additional data, but uses non-constrained computational resources for training, meaning the model is trained on 32 TPUs for 10 days.
The system presented in \cite{Baskar-2021-EatEnhancedASR-TT} has the weakest baseline, but shows the largest improvements by using a TTS system with x-vector \cite{8461375} speaker embeddings and the scaling of the synthetic encoder states. As seen in section \ref{sec:aed}, only changing the speaker embedding had a notable effect.
Investigating why the speaker embedding method is important although the encoder parts of ASR systems do not seem to benefit from synthetic data can be future work, together with finding objective performance markers for the quality of synthetic data different from UDR and WDR.

\section{Conclusion}

In this work we presented four state-of-the-art ASR systems for LibriSpeech-100h and tried to improve the performance of each by using synthetic data from a TTS system trained on the same data.
By using the alignment of a GMM-HMM system for silence removal, we were able to improve the stability of an autoregressive TTS system with respect to unaligned duration rate and word deletion rate.
We found that an increase in the stability of the TTS systems is not needed to generate useful synthetic data to be used in AED-ASR training.
For the first time we apply synthetic data from a TTS system to an attention-encoder-decoder, hybrid, CTC  and a RNN-T system in a direct comparison, and show that only the AED system can be significantly improved by using synthetic data. We show that we can get up to 38\% relative improvements by adding synthetic data to the AED-ASR system, even when using internal language model subtraction. This indicates that the benefit of adding synthetic data from additional text is mostly related to improving the robustness of the attention mechanism for sequence mapping, and not related to improving the internal language model of the decoder. Nevertheless, the improvement on AED systems is currently not sufficient to close the performance gap to a strong hybrid baseline presented in this work, which outperforms any other system in literature under the same data conditions with a word-error-rate of 3.3\%/10.0\% on test-clean and test-other respectively.

\vspace{-0.8em}
\section{Acknowledgements}
\vspace{-0.6em}
\footnotesize
This work has received funding from the European Research Council (ERC) under the European Union’s Horizon 2020 research
and innovation programme (grant agreement No 694537, project ”SEQCLAS”). This work was partly funded by the Google Focused Award "Pushing the Frontiers of ASR: Training Criteria and Semi-Supervised Learning".
The work reflects only the authors’ views and none of
the funding parties is responsible for any use that may be made of the information it contains. We like to thank Mattia Di Gangi, Christoph Lüscher, Peter Vieting and Wei Zhou for additional contributions.

\bibliographystyle{IEEEbib}
\nocite{*}
\bibliography{mybib}

\end{document}